\documentclass{Interspeech}

\usepackage{multirow}
\usepackage{tabularx}

\interspeechcameraready

\title{VARAN: Variational Inference for Self-Supervised Speech Models Fine-Tuning on Downstream Tasks}

\author[affiliation={1}]{Daria}{Diatlova}
\author[affiliation={2}]{Nikita}{Balagansky}
\author[affiliation={1}]{Alexander}{Varlamov}
\author[affiliation={1}]{Egor}{Spirin}

\affiliation{}{VK Lab}{Russia}
\affiliation{}{T-Tech}{Russia}
\email{n.n.balaganskiy@tbank.ru}

\usepackage{comment}
\usepackage{subcaption}

\newcommand{\varan}{\emph{VARAN}\ }

\begin{document}

\maketitle

\begin{abstract}
    
    Conventional methods for aggregating layers in fine-tuned self-supervised speech models, such as using the final layer or weighted sum, suffer from information bottlenecks and static feature weighting for all dataset examples. We propose VARAN, a framework that dynamically tailors layer aggregation to individual inputs. By employing layer-specialized probing heads and data-dependent weighting, VARAN adaptively prioritizes layer's features based on input. Evaluations on automatic speech recognition (ASR) and speech emotion recognition (SER) tasks demonstrate VARAN’s superior performance, particularly when using the LoRA fine-tuning technique. The framework resolves the trade-off between preserving layer-specific information and enabling flexible feature utilization, advancing efficient adaptation of self-supervised speech representations.
\end{abstract}

\section{Introduction}
Self-supervised learning (SSL) has revolutionized speech processing by enabling models like WavLM~\cite{chen2022wavlm} and Data2Vec~\cite{baevski2022data2vec} to learn rich, transferable representations from vast unlabeled audio corpora. These models excel in downstream tasks—such as ASR, SER, SV and others by leveraging hierarchical features across transformer encoder layers. However, effectively utilizing these features remains challenging. Conventional approaches aggregate layer-wise representations either by relying solely on the final layer or combining them via static, learnable weights. While simple, these methods suffer from two critical limitations: an \textit{information bottleneck}, where compressing multi-layer features into a single vector discards nuanced hierarchical information (e.g., phonetic details in lower layers vs. semantic context in higher ones), and \textit{static aggregation}, which applies fixed weights across all inputs, ignoring the variability in input complexity (e.g., ambiguous utterances requiring multi-layer analysis versus straightforward ones needing focused high-level features).

To address these issues, we propose \textbf{VARAN} (Variational Adaptive Layer Aggregation Network), a novel framework that dynamically tailors layer aggregation to individual inputs. \varan introduces two key modifications:
\begin{itemize}
    \item Layer-Specialized Probing Heads: Lightweight task-specific models are applied to each encoder layer, preserving distinct hierarchical features without cross-layer interference.
    \item Data-Dependent Weighting: A variational mechanism learns input-specific weights to prioritize layers, enabling the model to adaptively combine features based on input complexity.
\end{itemize}

Extensive experiments on ASR and SER tasks demonstrate \varan’s better or on-par results relative to static aggregation baselines, both in the standard fine-tuning setup and when applying LoRA.

\section{Prelimenaries}
\subsection{SSL Models' Architecture}
Modern SSL backbones share a similar architectural design that consists of two fundamental components: a feature extractor $\mathbf{F}$, typically implemented using CNNs, and a transformer encoder $\mathbf{E}$. The Transformer encoder is composed of $n$ layers which are denoted as $\mathbf{L}_i$. 

During the forward pass on downstream tasks, a raw waveform $\boldsymbol{x}$ first passes through the feature extractor to produce an initial representation $\boldsymbol{h}_0 = \mathbf{F}(\boldsymbol{x})$. Subsequently, $\boldsymbol{h}_0$ is processed through a series of transformer encoder layers. Each layer computes its output as $\boldsymbol{h}_i = \mathbf{L}_i(\boldsymbol{h}_{i-1})$.

After completing the forward pass, a sequence of hidden states $H = (\boldsymbol{h}_0, \dots, \boldsymbol{h}_n)$ is obtained. This set of hidden states is then used for downstream tasks.

\subsection{Layer Aggregation For Downstream Tasks}\label{sec:2_2}
The formulation of downstream tasks usually involves predicting $\boldsymbol{y}$ (i.e., class label, text, or speaker embedding) from waveform $\boldsymbol{x}$. This requires the SSL model, trained for the downstream task to have a probing head 
$p_\theta(\boldsymbol{y}|\boldsymbol{x})$ 
$ = \psi\left(\hat{\boldsymbol{h}}\right)$, 
where $\psi$ denotes the downstream model, 
$\theta$ are the model parameters
and $\hat{\boldsymbol{h}}$ are the hidden states of the model. In the simplest case $\hat{\boldsymbol{h}} = \boldsymbol{h}_n$, which we refer to as the \textit{last layer}. However, this approach does not utilize useful features from lower layers, leading to suboptimal performance. To address this issue, one can use learnable weights $w=(w_0, ..., w_n)$ to sum the representations from different layers

\begin{equation}\label{eq:weighted_sum}
    \hat{\boldsymbol{h}} = \sum_{i=0}^n w_i\cdot \boldsymbol{h}_i\ .
\end{equation}

\begin{figure*}[htbp]
    \centering
    \includegraphics[width=0.99\linewidth]{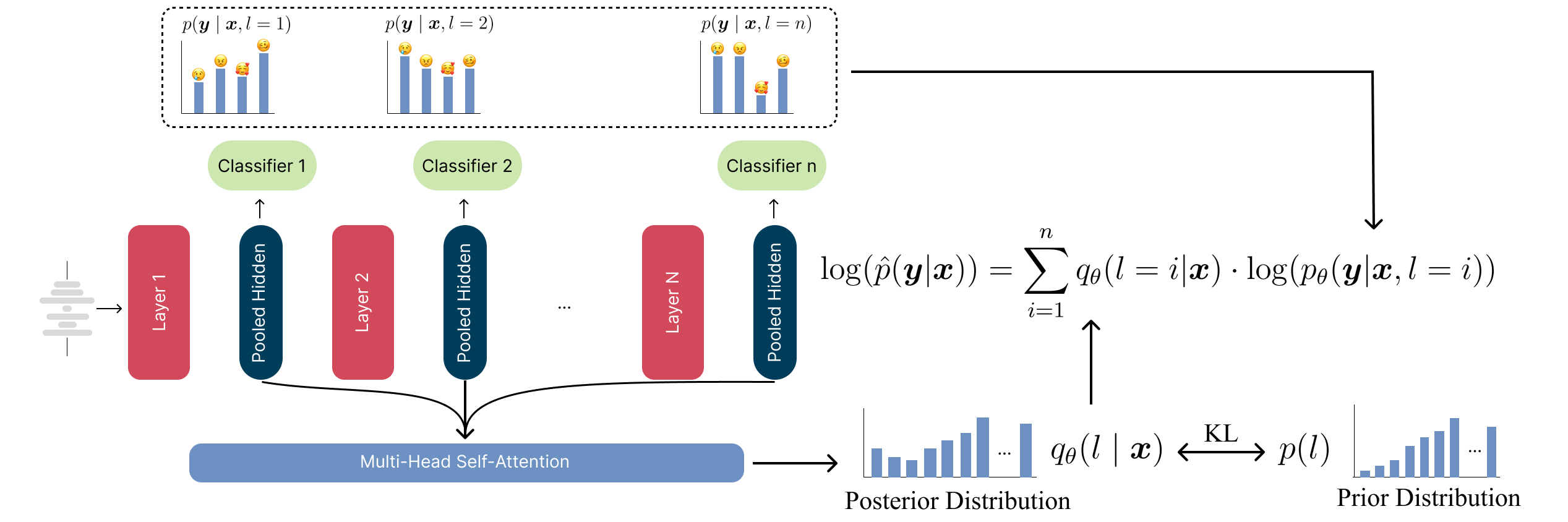}
    \caption{Overview of the VARAN layer aggregation method.}
    \label{fig:varan}
\end{figure*}

\subsection{Efficient Fine-Tuning Methods}

The simplest approach to adapt a pre-trained model for a downstream task is full fine-tuning, where the model is initialized with pre-trained weights and all parameters are updated according to the training data. However, this approach often leads to challenges such as catastrophic forgetting \cite{kemker2017measuring} and representation collapse, particularly when the downstream task has limited training data. To address these issues, parameter-efficient fine-tuning methods have emerged as a compelling alternative. These methods not only reduce the number of trainable parameters but also help retain the valuable knowledge acquired during pre-training.

One prominent example of such methods is LoRA \cite{hu2021lora}, which offers an elegant solution by freezing the original model parameters and introducing low-rank matrices. Specifically, instead of updating the entire weight matrix $W$, LoRA decomposes the update $\Delta W$ into the product of two smaller matrices:
\begin{equation}
    \Delta W = A \cdot B,
\end{equation}
where $A \in \mathbb{R}^{d \times r}$ and $B \in \mathbb{R}^{r \times k}$ are trainable matrices, and $r \ll \min(d, k)$ represents the rank of the decomposition. By constraining the updates to a low-rank subspace, LoRA significantly reduces the number of trainable parameters while preserving the pre-trained model's knowledge. This enables efficient adaptation to downstream tasks without compromising the quality of the learned representations.

\section{Method}\label{sec:method}
In this section, we propose VARAN, a Variational Inference-based method for aggregating self-supervised speech model outputs. We begin by outlining the limitations of current approaches. We then explain how our method addresses them, provide its theoretical foundation, derive the training objective, and finally detail the resulting architectural changes to the model.

\subsection{Limitations of Current Aggregation Methods}\label{limitations}

A common approach for aggregating hidden states before applying the probing head for a any downstream task, as described in Section~\ref{sec:2_2}, introduces two key limitations: (1) an \textit{information bottleneck} resulting from compressing layer-specific features into a single vector $\hat{\boldsymbol{h}}$, and (2) a \textit{static layer prioritization}, where a fixed set of weights $w$ ignores the optimal combination of features $H$ that depends on the input $\boldsymbol{x}$.

\begin{figure*}[t!]
    \centering
    \begin{subfigure}[b]{0.241\textwidth}
        \centering
        \includegraphics[height=1.2in]{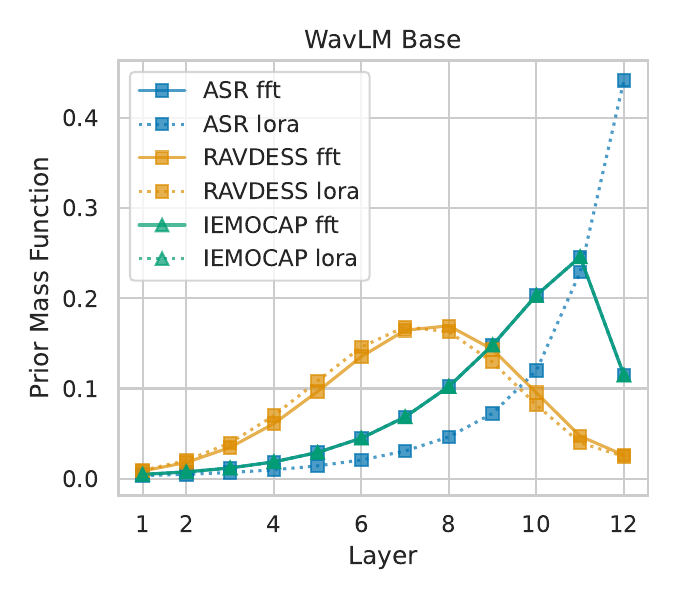}
    \end{subfigure}%
    ~ 
    \begin{subfigure}[b]{0.241\textwidth}
        \centering
        \includegraphics[height=1.2in]{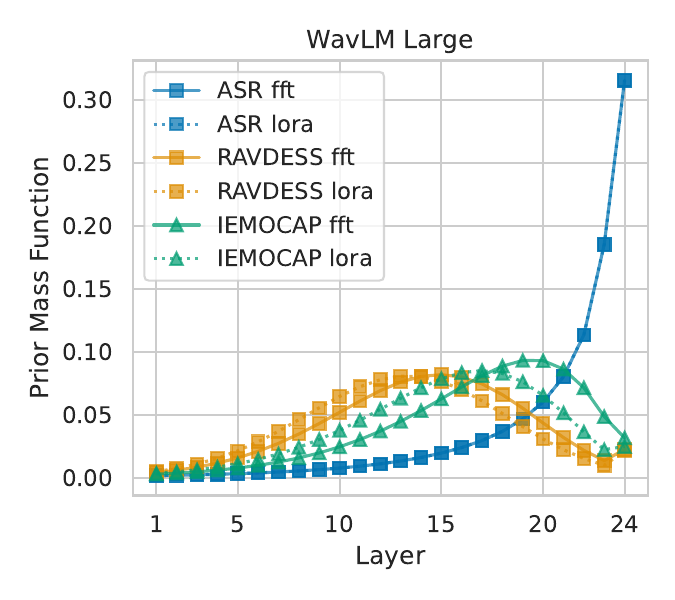}
    \end{subfigure}
    ~
    \begin{subfigure}[b]{0.241\textwidth}
        \centering
        \includegraphics[height=1.2in]{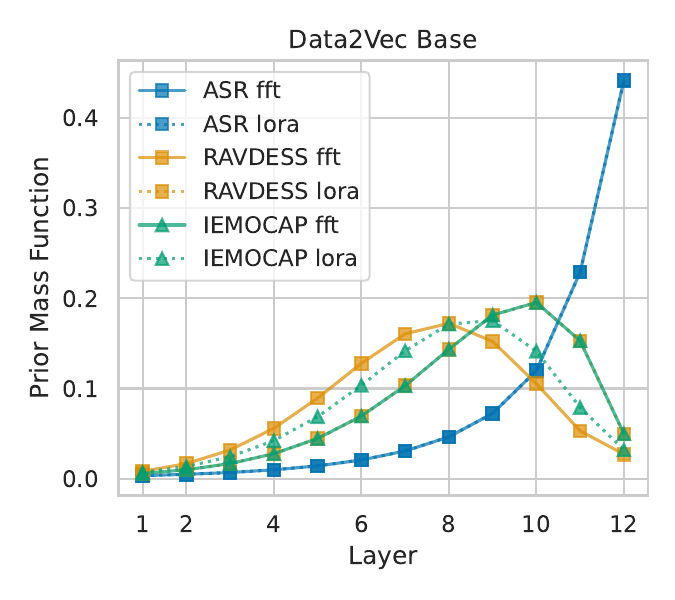}
    \end{subfigure}%
    ~ 
    \begin{subfigure}[b]{0.241\textwidth}
        \centering
        \includegraphics[height=1.2in]{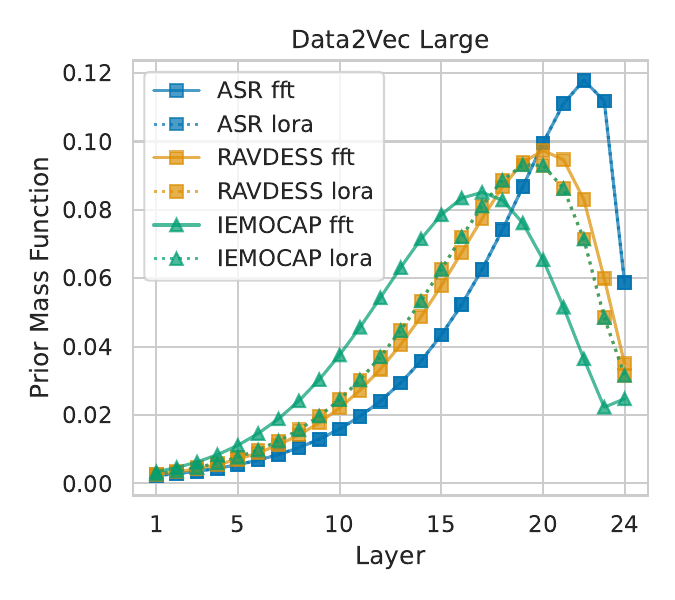}
    \end{subfigure}
    \caption{Prior distributions found during hyperparameters search. Models trained on RAVDESS\cite{livingstone2018ryerson} dataset tend to have distribution with mode on the middle layers. Models trained on ASR task (GigaSpeech \cite{DBLP:journals/corr/abs-2106-06909} dataset) in oposite tend to have distribution with mode on the last layer. See Section \ref{sec:analysis} for more details.} 
    \label{fig:prior}
\end{figure*}

\subsection{VARAN: Variational Inference Perspective}
To address the limitation of the existing layer aggregation methods we suggest to view a downstream task from variational inference perspective. We assume that for any downstream task, our goal is to predict the target $\boldsymbol{y}$ using input data $\boldsymbol{x}$. As mentioned in ~\ref{sec:2_2} one could use a simple linear classifier on top of the weighted sum of hidden layers, see Equation~\ref{eq:weighted_sum}, but we found this setup to be suboptimal, as it may require different layer weights $w$ to achieve the best performance for different tasks and even for individual samples. Choosing the right layer or combination of layers is challenging, as the training data does not have the optimal layer distribution.

However, we can learn it with the help of variational inference. In our setup, we have a classifier on top of each backbone's layer $p_\theta(\boldsymbol{y}\mid \boldsymbol{x}, l=i)$ and layer classifier $q_\theta(l \mid \boldsymbol{x})$, where $l$ is a discrete random variable and $i$ is a layer index. To maximize $p(\boldsymbol{y}\mid \boldsymbol{x})$, we can use the following objective:

\begin{align}
\label{eq:training_objective_vi}
L(\theta) &= - \mathbb{E}_{q_\theta(l \mid \boldsymbol{x})} \log p_\theta(\boldsymbol{y}\mid \boldsymbol{x}, l) \nonumber \\
&+ \operatorname{KL} \left(q_\theta(l\mid \boldsymbol{x}) \mid p(l)\right) 
\nonumber \\
&\geq - \log p(\boldsymbol{y}\mid \boldsymbol{x})
\end{align}

where $p(l)$ is a prior distribution on layer index. 

You can find more details on derivation in Appendix~\ref{appendix:elbo_deriviation}.

\subsection{VARAN: Training}
The training objective $L(\theta)$ derives from a variational upper bound, see Equation~\ref{eq:training_objective_vi}. Inspired by $\beta$-VAE \cite{higgins2017betavae}, we use a $\beta$ scaling factor as the regularization term. With this modification, our objective can be written as:
\cite{Kingma2013VAE}:  
\begin{align}
\label{eq:objective}
    L(\theta) &= - \mathbb{E}_{q_\theta(l \mid \boldsymbol{x})} \log p_\theta(\boldsymbol{y}\mid \boldsymbol{x}, l) \nonumber \\
    &+ \beta \operatorname{KL} \left(q_\theta(l\mid \boldsymbol{x}) \mid p(l)\right),
\end{align}
where the first term is a common downstream-task loss (eg. Cross Entropy for classification) computed for each $\psi_i(\boldsymbol{h}_i)$ and averaged with the predicted set of weighs $\{w_i(\boldsymbol{h}_i)\}_{i=1}^n$. 
Where $\psi_i$ is a probing head for each encoder layer $\mathbf{L}_i$.
Note that $\{w_i(\boldsymbol{h}_i)\}_{i=1}^n$ obtained individually for each sample $\boldsymbol{x}$.  Crucially, unlike Gumbel-based methods \cite{jang2016categorical}, gradients flow directly through $q_\theta(l|\boldsymbol{x})$, ensuring stable training of data-dependent weights.  

The second KL term regularizes the predicted distribution for each training sample toward a prior $p(l)$ defined for each downstream task. In practice, we used discretized reversed $\chi^2$ distribution and vary degrees of freedom during the hyperparameter search.

\subsection{VARAN: Inference}
At inference, predictions combine layer outputs via learned weights:  
\begin{align}\label{eq:inference}
    \log(\hat{p}(\boldsymbol{y}|\boldsymbol{x})) &= \sum_{i=1}^{n}q_\theta(l=i|\boldsymbol{x}) \cdot \log(p_\theta(\boldsymbol{y}|\boldsymbol{x}, l=i)) \nonumber \\  
    &= \sum_{i=1}^{n}w_i(\boldsymbol{h}_i) \cdot \log(\psi_i(\boldsymbol{h}_i)).  
\end{align}

VARAN addreses both an information botteleneck and a static layer prioritization issues: the first component of Equation~\ref{eq:inference} ensures that the model learns a unique set of weights $w_i$ for each sample to enable flexible, dynamic feature integration. The second component overcomes the information bottleneck by introducing different probing heads $\psi_i$ that extract specialized information from each layer, incorporating these rich representations into the final output. 

\subsection{VARAN: Architectural Perspective}

An overview of the proposed method can be found in Figure~\ref{fig:varan}. To compute $q_\theta(l=i|\boldsymbol{x})$ the posterior distribution predictor is parameterized using Multi-Head Self-Attention (MHSA). The input to the MHSA consists of hidden states from the Transformer Encoder blocks $H = (\boldsymbol{h}_1, \dots, \boldsymbol{h}_n)$, where $\mathbf{h} \in \mathbb{R}^{\text{b} \times \text{s} \times \text{d}}$ (with dimensions corresponding to batch size $b$, sequence length $s$, and hidden size $d$). We apply mean pooling along the sequence length dimension and concatenate the results into a single tensor. This tensor is fed into the standard MHSA mechanism, which acts as a feature mixer across layers. 
Finally, we apply a linear layer to the last dimension to obtain a tensor $\mathbf{H_\text{projected}} \in \mathbb{R}^{\text{b} \times \text{L} \times 1}$, perform a squeeze operation to remove the singleton dimension and apply softmax function along the layer dimension to produce weights $\mathbf{w} \in \mathbb{R}^{\text{b} \times \text{L}}$ for each batch sample. To get $\log(p_\theta(\boldsymbol{y}|\boldsymbol{x}, l=i))$ we employ separate lightweight models $\psi_i$ for each output of the encoder layer $\boldsymbol{h}_i$ which in our implementation is a simple linear layer. The final vector is obtained using the weighted sum as in Equation~\ref{eq:weighted_sum}. The detailed implementation of the proposed method can be found at \url{https://github.com/deepvk/varan}.

\begin{table}[hb]
    \centering
    \begin{tabular}{@{} l| c c @{}}
        \toprule
        \multirow{2}{.0\linewidth}{Parameter} & \multicolumn{2}{c}{Values} \\ 
        \cmidrule(lr){2-3}
         & ASR & SER \\
        \midrule
        LoRa Rank & \{16\} & \{8, 16, 32\} \\
        Batch Size & \{16\} & \{16, 32, 64\} \\
        Weight Decay & \{0.1\} & \{0.05, 0.1\} \\
        Learning Rate & \{1e-5, 1e-4, 1e-3\} & \{3e-5, 1e-5, 1e-4\} \\
        KL $\beta$ & \{0.05\} & \{0.01, 0.05, 0.1\} \\
        $\chi^2$ DF & \{1, 3\} & \{3, 5, 15, 35, 50, 100, 400\} \\
        \bottomrule
    \end{tabular}
    \caption{Hyperparameter search space.}
    \label{tab:hp}
\end{table}

\section{Experiments}
In this section, we describe the experimental setup and the results obtained by comparing the introduced \varan to existing methods and techniques for fine-tuning SSL models on downstream tasks.

\begin{table*}[t]
    \caption{
The results of \varan are compared with other layer aggregation methods across various fine-tuning techniques. \varan outperforms alternative methods in most experiments and achieves comparable performance in a few cases. For clarity, the best results within the same backbone and fine-tuning method are highlighted in \textbf{bold}, while comparable performances are \underline{underlined}.
}
    \centering
    \begin{tabular}{l|l|cc|cc|ccc}
        \toprule
    \multicolumn{2}{c|}{\multirow{2}{*}{\textbf{Method}}} & \multicolumn{4}{c|}{\textbf{Speech Emotion Recognition}} & \multicolumn{3}{c}{\textbf{Automatic Speech Recognition}} \\
    
    \multicolumn{2}{c|}{} & \multicolumn{2}{c|}{\textbf{RAVDESS~\cite{livingstone2018ryerson}}} & \multicolumn{2}{c|}{\textbf{IEMOCAP~\cite{busso2008iemocap}}} & \textbf{GigaSpeech~\cite{DBLP:journals/corr/abs-2106-06909}} & \multicolumn{2}{c}{\textbf{LibriSpeech}~\cite{panayotov2015librispeech}} \\
  
    \multicolumn{2}{c|}{} & \shortstack{Acc. ($\uparrow$)} & \shortstack{Weighted \\ F1 ($\uparrow$)} & \shortstack{Acc. ($\uparrow$)} & \shortstack{Weighted \\ F1 ($\uparrow$)} & WER ($\downarrow$) & \shortstack{test-clean, \\ WER ($\downarrow$)} & \shortstack{test-other, \\ WER ($\downarrow$)} \\
    
    \midrule
    \multicolumn{9}{c}{\textbf{Data2Vec Base~\cite{baevski2022data2vec}}} \\
    \midrule
        \multirow{3}{*}{Fine-Tuning}
            & Last Layer   & 0.50 & 0.46 & 0.61 & 0.55 & \textbf{33.53} & \textbf{14.90} & \textbf{22.18} \\
            & Weighted Sum & 0.49 & 0.47 & \underline{0.65} & 0.63 & 79.20 & 61.69 & 75.56 \\
            & \varan       & \textbf{0.57} & \textbf{0.56} & \underline{0.65} & \textbf{0.65} & 42.93 & 25.32 & 31.97 \\ \midrule
        \multirow{3}{*}{LoRA} 
            & Last Layer   & 0.48 & 0.42 & \underline{0.71} & 0.69 & 37.92 & 20.76 & 26.81 \\
            & Weighted Sum & 0.52 & 0.49 & \underline{0.71} & 0.69  & 42.39 & 24.09 & 31.16 \\
            & \varan       & \textbf{0.60} & \textbf{0.60} & \underline{0.71}  & \textbf{0.70} & \textbf{36.04} & \textbf{18.71} & \textbf{24.67} \\
    \midrule
    \multicolumn{9}{c}{\textbf{Data2Vec Large~\cite{baevski2022data2vec}}} \\
    \midrule
        \multirow{3}{*}{Fine-Tuning} 
            & Last Layer   & 0.39 & 0.34 & \textbf{0.71} & \textbf{0.67} & 27.70 & \textbf{12.01} & \textbf{16.24} \\
            & Weighted Sum & 0.73 & 0.71 & 0.66  & 0.64 & 63.31 & 44.75 & 60.00 \\
            & \varan       & \textbf{0.79} & \textbf{0.79} & 0.66 & 0.63 & \textbf{27.53} & 12.90 & 17.89\\ \midrule
        \multirow{3}{*}{LoRA} 
            & Last Layer   & 0.40 & 0.36 & 0.69 & 0.65 & \textbf{27.12} & \textbf{13.38} & \textbf{17.40} \\
            & Weighted Sum & 0.66 & 0.65 & 0.70 & 0.69 & 34.39 & 18.71 & 25.01 \\
            & \varan       & \textbf{0.72} & \textbf{0.72} & \textbf{0.71} & \textbf{0.71} & 27.79 & 13.97 & 18.03 \\
    \midrule
    \multicolumn{9}{c}{\textbf{WavLM Base~\cite{chen2022wavlm}}} \\
    \midrule
        \multirow{3}{*}{Fine-Tuning} 
            & Last Layer   & 0.65 & 0.64 & \textbf{0.68} & \textbf{0.68}  & \textbf{50.73} & \textbf{31.92} & \textbf{44.19} \\
            & Weighted Sum & 0.64 & 0.64 & 0.65  & 0.62 & 67.35 & 49.33 & 65.46 \\
            & \varan       & \textbf{0.70} & \textbf{0.70} & 0.66 & 0.63 & 59.66 & 41.11 & 54.89 \\ \midrule
        \multirow{3}{*}{LoRA} 
            & Last Layer   & 0.48 & 0.48 & 0.70  & 0.68 & 48.70 & \textbf{29.20} & 39.94 \\
            & Weighted Sum & 0.54 & 0.54 & 0.66 & 0.65 & 52.56 & 33.60 & 44.00 \\
            & \varan       & \textbf{0.65} & \textbf{0.65} & \textbf{0.72} & \textbf{0.71} & \textbf{47.83} & 29.25 & \textbf{39.19} \\
    \midrule
    \multicolumn{9}{c}{\textbf{WavLM Large~\cite{chen2022wavlm}}} \\
    \midrule
        \multirow{3}{*}{Fine-Tuning}
            & Last Layer   & 0.46 & 0.42 & 0.69 & 0.68 & 31.62 & 18.59 & 26.34 \\
            & Weighted Sum & 0.73 & 0.71 & \underline{0.73} & \textbf{0.73} & 26.70 & \textbf{14.96} & \textbf{20.30} \\
            & \varan       & \textbf{0.75} & \textbf{0.76} & \underline{0.73} & 0.72 & \textbf{26.29} & 15.39 & 20.55 \\ \midrule
        \multirow{3}{*}{LoRA}
            & Last Layer   & 0.51 & 0.50 & 0.72 & 0.71 & 32.92 & 20.94 & 26.07 \\
            & Weighted Sum & \underline{0.60} & \underline{0.56} & 0.72 & 0.72 & 32.32 & 18.38 & 25.31 \\
            & \varan       & \underline{0.60} & \underline{0.56} & \textbf{0.73} & \textbf{0.73} & \textbf{27.39} & \textbf{15.55} & \textbf{21.17} \\
        \bottomrule
    \end{tabular}
    \label{table:main}
\end{table*}

\subsection{Setup}

To examine the \varan, we selected the Automatic Speech Recognition (ASR) and Speech Emotion Recognition (SER) tasks. These tasks require an understanding of acoustic features at both the semantic and phonetic levels, as well as speaker-dependent information. For the ASR task, we follow the experimental setup described in~\cite{zaiem2024less}. Specifically, we use the XS subset of the GigaSpeech dataset~\cite{DBLP:journals/corr/abs-2106-06909} for training. In addition to the test portion of the GigaSpeech dataset, we report results on the "test clean" and "test other" sets from the LibriSpeech dataset~\cite{panayotov2015librispeech}. For the SER task, we conduct experiments using the RAVDESS~\cite{livingstone2018ryerson} and IEMOCAP~\cite{busso2008iemocap} datasets. For the RAVDESS dataset, we use the original train-validation-test split and merge the neutral emotion with calm, resulting in a total of 7 training labels. For the IEMOCAP dataset, we perform 10-fold cross-validation by training on 9 speakers and testing on the remaining one.

As baselines for layer aggregation methods, we employ two approaches: (1) the straightforward \textit{last layer} method, where no aggregation occurs and features from the final layer are directly used; and (2) the \textit{weighted sum} method, where features are aggregated using learnable weights without conditioning on sample variation. Furthermore, we investigate different fine-tuning techniques. Initially, we keep the CNN feature extractor frozen and train the rest of the model's parameters on a downstream task, which can be effective but computationally expensive, we refer this setup as \textit{Fine-Tuning}. We also apply parameter-efficient fine-tuning using LoRA~\cite{hu2021lora}, a method that is both fast and stable for training large SSL models on downstream tasks, particularly under limited computational resources. More specifically, we follow~\cite{zaiem2024less} by replacing the feed-forward layers that come after the self-attention mechanism with LoRA layers.

For the backbones, we use both WavLM~\cite{chen2022wavlm} and Data2Vec~\cite{baevski2022data2vec} in base and large configurations. For both ASR and SER tasks, we used 2 fully-connected layers with a hidden size of 256 as a probing head. ASR decoding was conducted at the character level using beam search (beam size=5, beam threshold=20), with no language model integration.

Since \varan’s training objective derives from a variational upper bound (see Section~\ref{sec:method}), one essential consideration is selecting an appropriate prior distribution. We adopt a discretized reversed $\chi^2$ as the prior, with degrees of freedom treated as hyperparameters during search.

For each fine-tuning, we use grid search to find the optimal hyperparameters on the validation sets of the corresponding datasets. See Table \ref{tab:hp} for hyperparameter grid. For the IEMOCAP dataset, "Session 5" is used. Full grid of the hyperparameters can be found in Table \ref{tab:hp}.

\subsection{Results}
\label{sec:4_2}

Experimental results are summarized in Table~\ref{table:main}. On speech emotion recognition, \varan consistently outperforms other layer aggregation methods across configurations. The sole exception occurs in the LoRA fine-tuning setup with WavLM Large, where \varan achieves performance comparable to weighted-sum aggregation.

For the IEMOCAP dataset, \varan surpasses conventional methods less frequently in \textit{Fine-Tuning} setup. However, the strongest results on this benchmark are observed in the LoRA fine-tuning regime, where \varan delivers substantial improvements over all baselines, achieving a relative accuracy gain of 2.87\% compared to weighted-sum aggregation. 

According to evaluations on the GigaSpeech ASR corpus and LibiSpeech test-other subsets, \varan outperforms other methods in the LoRA fine-tuning setup across all models except Data2Vec Large. The relative improvement in WER on the GigaSpeech test set is 7.7\% compared to the last-layer baseline. In the \textit{Fine-Tuning} setup, performance gains are less stable: the last layer predominantly outperforms other methods.

This phenomenon can be explained by two observations. First, ~\cite{toyota23} demonstrated that models trained to predict discrete units learn phonetic and word information concentrated in higher layers. Second, during LoRA fine-tuning, models retain more pretraining knowledge~\cite{zaiem2024less}, making layer aggregation in a data-dependent setup beneficial. Conversely, in \textit{Fine-Tuning}, knowledge distribution may collapse to the last layer, diminishing the utility of multi-layer aggregation.

In general, \varan consistently outperforms other methods in most scenarios, particularly in the LoRA setup and the speech emotion recognition task, where low-level features and layer weighting play a crucial role in final performance.

\subsection{Analysis}
\label{sec:analysis}

To investigate the differences between datasets and models we plot the PMF function of prior distributions $q(l)$ found during hyperparameter search for \varan models. The results are presented in Figure \ref{fig:prior}. The first observation is that both Data2Vec \cite{baevski2022data2vec} base and large models benefit from the weights skewed towards the last layers in contrast to WavLM \cite{chen2022wavlm}, which indicates that features needed for the tasks are grouped there. The second observation is that ASR benefits most from the last hidden state.

\section{Related Work}
\subsection{Speech Self-Supervised Model Fine-Tuning}
Recent work has explored strategies to optimize speech SSL models for downstream tasks through architectural and training adaptations. \cite{zaiem2024speech} demonstrated that scaling probing head size not only improves model-specific performance on SER and SV tasks but also reshuffles SSL model rankings in benchmark evaluations. Work by \cite{zaiem2024less} investigated continual learning in SSL fine-tuning for ASR, revealing that full model unfreezing induces pre-training knowledge forgetting, whereas parameter-efficient methods like LoRA and Elastic Weight Consolidation mitigate this issue and achieve superior word error rates (WER), particularly on out-of-distribution data. Complementary to these approaches, \cite{peng2022ieee} proposed Multi-Head-Factorized Attentive Pooling (MHFA) for SV to hierarchically aggregate low-level speaker features with high-level phonetic information, thereby enhancing feature disentanglement.

\subsection{Layer Analysis of Speech Self-Supervised Models}
\cite{chen2022wavlm} found that layers at the beginning and middle of the WavLM model had higher magnitudes of learned weights after fine-tuning on speaker-dependent tasks, indicating their importance for speaker verification (SV) and speaker identification (SI). \cite{toyota23} used CCA analysis to show that models trained to predict discrete units learn phonetic and word information in intermediate layers, with this information concentrated in higher layers. They also demonstrated that layer-wise phone and word content correlate well with downstream task performance for phoneme recognition (PR) and automatic speech recognition (ASR). \cite{Yang24Foundations} showed that weights learned via weighted sum are not correlated with layers' performance on tasks. Both \cite{Yang24Foundations} and \cite{toyota23} found that for some tasks and SSL models, the best individual layer outperforms the learned weighted sum. However, this improvement in single-layer benchmarking is specific to individual SSL models and is not consistent across all models~\cite{toyota23}.

\subsection{Variational Inference and Early Exit}

From one perspective, our method could be seen as VAE \cite{vae_kingma} with layer index $l$ as a latent variable. Therefore, we can successfully adapt some techniques for it, like \cite{higgins2017betavae}. \cite{banino2021pondernet} used a similar approach, but for adaptive computational time \cite{graves2016adaptive} inference. This method was also adapted for early exiting during inference \cite{palbert}. While reducing inference time of the models is a promising direction, in our work, we concentrated solely on performance improvement.

\section{Conclusion}
In our work, we present a novel layer aggregation method called VARAN, which uses per-sample weighting of the predictions from probing heads to obtain final predictions. Through the extensive experiments, we show that \varan method achieves better or comparable results on the SER and ASR tasks. We also study its performance across different models and setups, allowing as to analyze the importance of the layers for the different tasks. We believe our findings can amplify further research on feature aggregation methods.

\bibliographystyle{IEEEtran}
\bibliography{mybib}
\appendix
\section{ELBO}
\label{appendix:elbo_deriviation}

\newcommand{\Expec}{\mathbb{E}_{q}}

Our goal is to maximize the likelihood $p(\boldsymbol{y} \mid \boldsymbol{x})$. As SSL models use layer-aggregation methods for hidden representations from multiple layers, we need to derive a lower bound for the prior downstream task layer distribution.

For simplicity, let us denote the approximated posterior as $q_\theta(l\mid \boldsymbol{x})$, the true posterior which is a particular downstream task distribution as $p(l\mid \boldsymbol{x}, y)$, the prior layer distribution as  $p(l)$ and $\mathbb{E}_{l \sim q_\theta(l \mid \boldsymbol{x})}$ as $\Expec$~. Our goal is to approximate the posterior distribution so that it is maximally close to the true posterior. For that, let us derive a formula for KL-divergence:

\begin{equation*}
\begin{aligned}
    & \operatorname{KL}\left[ q_\theta(l \mid \boldsymbol{x}) \, \Vert \, p(l \mid \boldsymbol{x}, \boldsymbol{y})\right] \\
    &= \mathbb{E} \log \frac{q_\theta(l \mid \boldsymbol{x})}{p(l \mid \boldsymbol{x}, \boldsymbol{y})} \\
    &= \mathbb{E} \log \frac{q_\theta(l \mid \boldsymbol{x}) \cdot p(\boldsymbol{x}, \boldsymbol{y})}{p(l, \boldsymbol{y}, \boldsymbol{x})} \\
    &= \mathbb{E} \log \frac{q_\theta(l \mid \boldsymbol{x}) \cdot p(\boldsymbol{y} \mid \boldsymbol{x}) \cdot p(\boldsymbol{x})}{p(l, \boldsymbol{y} \mid \boldsymbol{x}) \cdot p(\boldsymbol{x})} \\
    &= \mathbb{E} \log \frac{q_\theta(l \mid \boldsymbol{x}) \cdot p(\boldsymbol{y} \mid \boldsymbol{x})}{p(l, \boldsymbol{y} \mid \boldsymbol{x})} \\
    &= \mathbb{E} \log \frac{q_\theta(l \mid \boldsymbol{x})}{p(l, \boldsymbol{y} \mid \boldsymbol{x})} + \mathbb{E} \log p(\boldsymbol{y} \mid \boldsymbol{x}) \\
    &= \mathbb{E} \log \frac{q_\theta(l \mid \boldsymbol{x})}{p(l, \boldsymbol{y} \mid \boldsymbol{x})} + \log p(\boldsymbol{y} \mid \boldsymbol{x}).
\end{aligned}
\end{equation*}

At this point, we still don't know how to access $p(l, \boldsymbol{y} \mid \boldsymbol{x})$. Let us rearrange the terms and write the log-likelihood $\log p(\boldsymbol{y} | \boldsymbol{x})$ as:

\begin{equation*}
\begin{aligned}
    \log p(\boldsymbol{y} \mid \boldsymbol{x}) &= \mathbb{E} \log \frac{p(l, \boldsymbol{y} \mid \boldsymbol{x})}{q_\theta(l \mid \boldsymbol{x})} + \operatorname{KL} \left(q_\theta(l \mid \boldsymbol{x}) \, \Vert \, p(l \mid \boldsymbol{x}, \boldsymbol{y})\right).
\end{aligned}
\end{equation*}

As log-likelihood is the logarithm of the probability, it is $\geq 0$. KL-divergence is a measure of distance which is also $\geq 0$, then the lower bound for likelihood can be written as:

\begin{equation*}
    \log p(\boldsymbol{y} \mid \boldsymbol{x}) \geq LB= \mathbb{E} \log \frac{p(l, \boldsymbol{y} \mid \boldsymbol{x})}{q_\theta(l
    \mid \boldsymbol{x})}
\end{equation*}

Our initial goal was to find a training objective for likelihood maximization. The training objective is the negative log-likelihood $-L(\theta) = LB$. Let us further simplify LB to bring it to a form which can be used to update gradients:

\begin{equation*}
\begin{aligned}
    LB &= -L(\theta) = \mathbb{E} \log \frac{p(l, \boldsymbol{y} \mid \boldsymbol{x})}{q_\theta(l \mid \boldsymbol{x})} = \mathbb{E} \log \frac{p(\boldsymbol{y} \mid \boldsymbol{x}, l) \cdot p(l)}{q_\theta(l \mid\boldsymbol{x})} \\
    &= \mathbb{E} \log p(\boldsymbol{y} \mid \boldsymbol{x}, l) + \mathbb{E} \log \frac{p(l)}{q_\theta(l \mid\boldsymbol{x})}.
\end{aligned}
\label{eq:lower_bound}
\end{equation*}

From the definition of KL-divergence, $\operatorname{KL}\left[q_\theta(l \mid \boldsymbol{x}) \, \Vert \, p(l)\right]=\mathbb{E} \log \frac{q_\theta(l \mid \boldsymbol{x})}{p(l)}$, we get that\\
$\mathbb{E} \log \frac{p(l)}{q_\theta(l \mid \boldsymbol{x})}=-\operatorname{KL}\left(q_\theta(l \mid \boldsymbol{x}) \, \Vert \, p(l)\right)$. Which gives us equation~\ref{eq:elbo-pre-final-form} that is the same as equation~\ref{eq:objective}:

\begin{equation}
\label{eq:elbo-pre-final-form}
    L(\theta)=-\mathbb{E} \log p(\boldsymbol{y} \mid \boldsymbol{x}, l) + \operatorname{KL}\left(q_\theta(l \mid \boldsymbol{x}) \, \Vert \, p(l)\right).
\end{equation}

By expanding $\mathbb{E}$ and selecting the downstream model, $p_\theta$, we get equation~\ref{eq:elbo-final-form}:

\begin{equation}
\begin{aligned}
\label{eq:elbo-final-form}
    L(\theta) &= -\sum_{i=1}^n q_\theta(l=i \mid \boldsymbol{x}) \cdot \log p_\theta(\boldsymbol{y} \mid l=i, \boldsymbol{x}) \\ 
    &+ \operatorname{KL}\left(q_\theta(l \mid \boldsymbol{x}) \, \Vert \, p(l)\right).
\end{aligned}
\end{equation}

\end{document}